\documentclass[conference]{IEEEtran}
\IEEEoverridecommandlockouts
\usepackage{cite}
\usepackage{amsmath,amssymb,amsfonts}
\usepackage[absolute,overlay]{textpos}

\usepackage{hyperref}
\usepackage{newtxmath} % for Times Roman clone math font
\usepackage[scr=rsfso]{mathalfa} % "oblique" version of mathrsfs
\usepackage{comment}
\usepackage[letterpaper, top=0.75in, left=0.62in, right=0.64in, bottom=1.1in]{geometry}
\usepackage{algorithmic}
\usepackage{graphicx}
\usepackage{textcomp}
\usepackage{xcolor}
\def\BibTeX{{\rm B\kern-.05em{\sc i\kern-.025em b}\kern-.08em
    T\kern-.1667em\lower.7ex\hbox{E}\kern-.125emX}}
\begin{document}

\title{Compound Fault Diagnosis for Train Transmission Systems Using Deep Learning with Fourier-enhanced Representation\\
%{\footnotesize \textsuperscript{*}Note: Sub-titles are not captured in Xplore and
%should not be used}
%\thanks{Identify applicable funding agency here. If none, delete this.}
}

\author{\IEEEauthorblockN{Jonathan Adam Rico}
\IEEEauthorblockA{\textit{Engineering Product Development} \\
\textit{SUTD}\\
\textit{Institute for Infocomm Research}\\
\textit{A*STAR}\\
Singapore \\
jonathanadam\_rico@mymail.sutd.edu.sg}
\and
\IEEEauthorblockN{Nagarajan Raghavan*}
\IEEEauthorblockA{\textit{Engineering Product Development} \\
\textit{SUTD}\\
Singapore \\
nagarajan@sutd.edu.sg}
*Corresponding author
\and
\IEEEauthorblockN{Senthilnath Jayavelu}
\IEEEauthorblockA{\textit{Institute for Infocomm Research} \\
\textit{A*STAR}\\
Singapore \\
j\_senthilnath@i2r.a-star.edu.sg}
}

\maketitle

\begin{textblock*}{\textwidth}(1.5cm,26cm) % Adjust coordinates (x,y) as needed
\footnotesize
\noindent
© 2025 IEEE. Personal use of this material is permitted. Permission
from IEEE must be obtained for all other uses, in any current or future
media, including reprinting/republishing this material for advertising or
promotional purposes, creating new collective works, for resale or
redistribution to servers or lists, or reuse of any copyrighted
component of this work in other works. \\
The final version is published in \textit{IEEE International Conference on Prognostics and Health Management (ICPHM)} and is available at: \url{https://doi.org/10.1109/ICPHM65385.2025.11062058}
\end{textblock*}

\begin{abstract}
Fault diagnosis prevents train disruptions by ensuring the stability and reliability of their transmission systems. Data-driven fault diagnosis models have several advantages over traditional methods in terms of dealing with non-linearity, adaptability, scalability, and automation. However, existing data-driven models are trained on separate transmission components and only consider single faults due to the limitations of existing datasets. These models will perform worse in scenarios where components operate with each other at the same time, affecting each component's vibration signals. To address some of these challenges, we propose a frequency domain representation and a 1-dimensional convolutional neural network for compound fault diagnosis and applied it on the PHM Beijing 2024 dataset, which includes 21 sensor channels, 17 single faults, and 42 compound faults from 4 interacting components, that is, motor, gearbox, left axle box, and right axle box. Our proposed model achieved 97.67\% and 93.93\% accuracies on the test set with 17 single faults and on the test set with 42 compound faults, respectively.

\end{abstract}


\section{Introduction}

Fault diagnosis plays a crucial role in maintaining the stability and reliability of transmission components, helping to prevent disruptions in train operations. Identifying and addressing faults early ensures that trains can run smoothly without unexpected interruptions, ultimately enhancing the overall safety and efficiency of the transportation system. Existing fault diagnosis models are trained on datasets that have motor current and vibration data on individual components \cite{ding2024_dataset}. In practice, train transmission systems are made up of several components such as motor, gearbox, and axle boxes that interact with each other. Moreover, multiple faults of the same or different components may occur at the same time.

Fault diagnosis using motor current and vibration signals is a time series classification task. Thus, time series signal processing methods, techniques, and representations can be applied such as spectral analysis \cite{antoni2007_spectral}, 
statistical and probabilistic analysis \cite{abid2018_statistical},
blind deconvolution \cite{buzzoni2018_deconvolution}, 
fast Fourier transform (FFT) \cite{mahgoun2013_fft},
short-time Fourier transform (STFT) \cite{geraei2024_stft},
and wavelet transform (WT) \cite{abad2016_wavelet}.
%statistical and probabilistic analysis (Xin et al. 2022) \cite{}, 
%and matrix analysis (Wang et al. 2022) \cite{}. 
The WT representation provides both spectral and temporal features, but given a static dataset, FFT representation can be enough. The FFT representation is widely utilized in different fault diagnosis fields \cite{mahgoun2013_fft}\cite{saribulut2013_fft} due to its ability to detect fault frequencies, identify multiple faults in a single analysis, and offer computational efficiency.

Several machine learning models were developed for fault diagnosis, such as the Support Vector Machine (SVM) \cite{yan2018_svm}, K-Nearest Neighbors
(KNN) \cite{sharma2018_knn}, 
Random Forest (RF) \cite{yan2019_rf}, 
and Extreme Learning Machine (ELM)  \cite{he2021_elm}. However, such models rely on feature extraction techniques to transform the input signals into structured tabular data. In contrast to machine learning models, deep learning models do not require feature extraction from time series vibration signals into structured tabular features. Some of the most commonly used deep learning models for fault diagnosis include recurrent architecture \cite{gao2021_rnn}, CNN \cite{li2023_cnn}, graph-based models \cite{ding2024_dataset}, autoencoder-based architectures \cite{wu2021_ae}, and transformer-based architectures \cite{luo2024_ffttrans}. A summary table of the commonly used methods for bearing fault diagnosis is shown in Table \ref{tab:existing_methods}.

\begin{table}[tb]
\caption{Existing methods for bearing fault diagnosis.}
\centering
\label{tab:existing_methods}
\begin{tabular}{llc}
\hline
Method & Algorithm & Related Literature \\ \hline
Traditional/ & Spectral analysis & \cite{antoni2007_spectral}  \\
Feature & FFT & \cite{mahgoun2013_fft} \\
Representation & WT & \cite{abad2016_wavelet} \\
& Blind Deconvolution & \cite{buzzoni2018_deconvolution} \\
& Statistical & \cite{abid2018_statistical} \\
& STFT& \cite{geraei2024_stft} \\ \hline
Machine Learning & SVM & \cite{yan2018_svm} \\
& KNN & \cite{sharma2018_knn} \\
& RF & \cite{yan2019_rf} \\
& ELM & \cite{he2021_elm} \\ \hline
Deep Learning & RNN & \cite{gao2021_rnn}\cite{pan2018_1dcnnlstm} \\
& CNN & \cite{ding2024_dataset} \cite{li2023_cnn} \cite{aldeoes2024_cwtcnn} \cite{wang2022_sscnn}  \cite{pan2018_1dcnnlstm} \cite{zhang2023_fft1dcnn} \\
& Autoencoder & \cite{wu2021_ae} \\
%& GAN & \cite{zhong2023_gan} \\ 
& GNN & \cite{ding2024_dataset} \\ 
& Transformer & \cite{luo2024_ffttrans} \\
\hline
\end{tabular}
\end{table}

The key contributions of our work are as follows:
\begin{itemize}
    \item We developed a fault diagnosis model that can detect compound faults in a train transmission system where the vibration measurements of each component affect the vibration measurements of adjacent component.
    \item By taking the amplitude spectrum of the raw signals with FFT and efficiently identifying the speed working condition for data normalization, we enhance the feature representation of the motor current and vibration signals. 
    \item Using 1DCNN feature extraction further improves the fault diagnosis due to the translational invariance of CNN making the deep learning model robust to frequency shifts.
\end{itemize}

\begin{comment}
    The succeeding sections of this paper are structured as follows: 
Chapter 2 discusses all the relevant studies and positions our work in the field of fault diagnosis.
Chapter 3 discusses the methodology preliminaries such as theoretical background of Fourier transform, Convolutional Neural Networks (CNN), and Supervised Autoencoder; deep learning subsection discusses the proposed model architecture and training process. Chapter 4 discusses the results including the data acquisition and challenges in the data, implementation details, the results of the benchmark experiments and the results of the ablation experiments. Chapter 5 summarizes the key takeaways of this paper, limitations, and future works.
\end{comment}

\section{Related Works}

%\subsection{Bearing Fault Diagnosis}

\begin{comment}
Existing fault diagnosis models were developed for separate components such as for wheelset bearing faults \cite{kou2020_wheelset} \cite{liu2021_wheelset}, gearbox faults \cite{wang2022_gearbox}, traction motor bearing faults \cite{zou2021_motor}, and both wheelset bearing and traction motor bearing faults \cite{wang2022_sscnn}.
\end{comment}

\begin{figure}
    \vspace{0.05in}
    \centering
    \includegraphics[width=0.9\columnwidth]{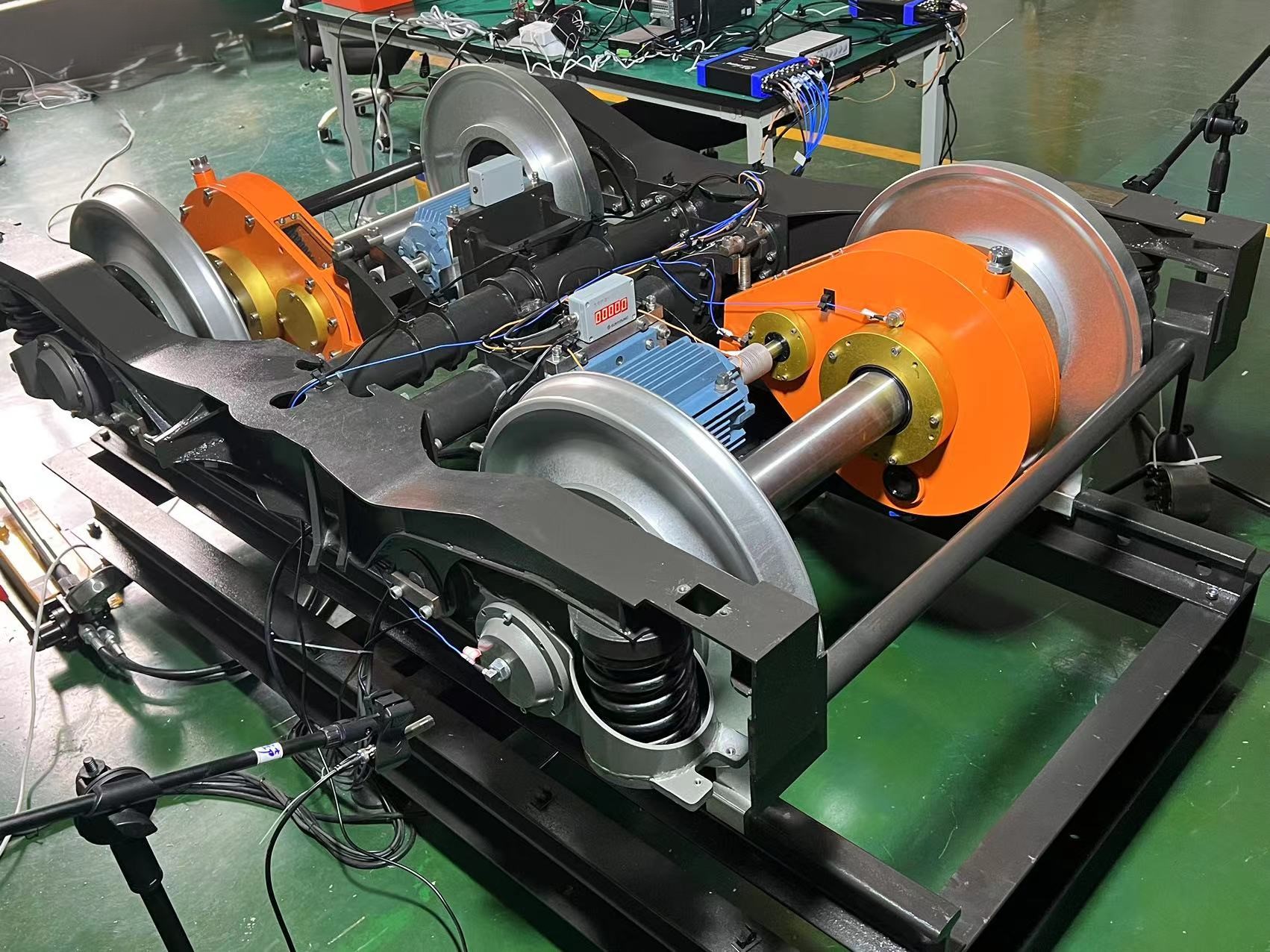}
    \caption{Photo \cite{ding2024_dataset} of the experimental platform of the train bogie used in the PHM Beijing 2024 dataset.}
    \label{fig:experimental_platform}
\end{figure}

Table \ref{tab:existing_datasets} shows the description of the existing bearing fault diagnosis datasets used in the literature. Datasets for bearing fault diagnosis, such as the Case Western Research University (CWRU) dataset \cite{cwru2009_dataset}, Paderborn University bearing dataset \cite{lessmeier2016_dataset}, Vishwakarma Institute of Technology (VIT) College motor bearing dataset \cite{aldeoes2024_cwtcnn}, High-speed train (HST) wheelset bearing dataset \cite{wang2022_sscnn},  Hanoi University of Science and Technology (HUST) bearing dataset \cite{hong2023_dataset}, and the HST axle box bearing dataset \cite{zhang2025_dataset} do not consider the interactions between components. Moreover, these datasets only tackle single faults in bearings. On the other hand, Prognostics and Health Management (PHM) Beijing 2024 dataset \cite{ding2024_dataset} consists of compound faults and considers vibrational interactions between transmission components. Fig. \ref{fig:experimental_platform} shows the experimental platform of the train bogie used to collect sensor data from the motor, gearbox, left axle box, and right axle box.

\begin{table*}[tb]
\caption{Description of bearing fault diagnosis datasets and the literature that used the datasets for fault diagnosis.}
\centering
\label{tab:existing_datasets}
\begin{tabular}{llllc}
\hline
Dataset & Date & Components & Fault Types & Related Literature \\ \hline
CWRU bearing dataset \cite{cwru2009_dataset} & 2009 & Bearings in motor &  9 single faults, 1 normal & \cite{gao2021_rnn}\cite{li2023_cnn}\cite{wu2021_ae}\cite{aldeoes2024_cwtcnn}\cite{pan2018_1dcnnlstm}\\
& & & & \cite{kou2020_wheelset}
\cite{liu2021_wheelset}\cite{wang2022_gearbox}\cite{zhang2023_fft1dcnn} \\
Paderborn University bearing dataset \cite{lessmeier2016_dataset} & 2016 & Bearings in rolling element & 26 single faults, 6 normal & \cite{li2023_cnn}\\ 
VIT College motor bearing dataset \cite{aldeoes2024_cwtcnn} & 2021 & Bearings in motor and gearbox & 2 single faults, 1 normal & \cite{aldeoes2024_cwtcnn}\\
HST wheelset bearing dataset \cite{wang2022_sscnn} & 2022 & Bearings in train axlebox & 10 single faults, 1 normal & \cite{wang2022_gearbox}\\
HUST bearing dataset \cite{hong2023_dataset} & 2023 & Bearings in rolling element & 6 single faults, 1 normal & \cite{abbasi2025_hust}\\
PHM Beijing dataset \cite{ding2024_dataset} & 2024 & Bearings in train motor, gearbox, & \bf{42 compound faults} & \cite{ding2024_dataset}\\
& & and axle boxes \\
HST axle box bearing dataset \cite{zhang2025_dataset} & 2025 & Bearings in train axle box & 8 single faults, 1 normal & \cite{zhang2025_dataset}\\ \hline
\end{tabular}
\end{table*}

%\subsection{Deep Learning for Fault Diagnosis}

\begin{comment}
    Data-driven fault diagnosis methods suffer from data imbalance and limited data availability. To address such issues, \cite{wu2021_ae} used Autoencoders so that the model only needs to train on the normal (majority) samples. Likewise, \cite{zhong2023_gan} used Generative Adversarial Networks (GANs) to generate synthetic samples for the anomaly (minority) class. 
\end{comment}

\begin{comment}
Related works that used FFT or 1DCNN for fault diagnosis include FFT-1DCNN \cite{zhang2023_fft1dcnn}, CWT-CNN \cite{aldeoes2024_cwtcnn},
SelfSupervisedCNN \cite{wang2022_sscnn},
FFT-Trans \cite{luo2024_ffttrans},
%FFT-Denoising AE (Lee 2024) \cite{} dissertation
and 1DCNN-LSTM \cite{pan2018_1dcnnlstm}. 
\end{comment}

% Raw signals
Numerous fault diagnosis models were proposed that utilize CNN architecture, including an adaptive multiscale fully convolutional network (AMFCN) \cite{li2023_cnn}, self-supervised CNN (SSCNN) \cite{wang2022_sscnn}, 1DCNN-LSTM\cite{pan2018_1dcnnlstm}, end-to-end CNN
\cite{kou2020_wheelset}, and multitask 1DCNN
\cite{liu2021_wheelset}. However, these models do not have data representation and directly input the raw signals into their CNN architectures. In addition, they develop fault diagnosis models for individual components and only tackle single faults. Recent studies in fault diagnosis incorporated feature representations in their deep learning models such as discrete wavelet transform (DWT) in self-paced CNN 
\cite{zhang2025_dataset}, CNN-based model \cite{wang2022_gearbox}, deep belief network (DBN) \cite{zou2021_motor}; continuous wavelet transform (CWT) in convolution feature-based RNN (CFRNN) \cite{gao2021_rnn}, CWT-CNN
\cite{aldeoes2024_cwtcnn}, CNN-LSTM \cite{abbasi2025_hust}; STFT representation in \cite{wu2021_ae} proposed hybrid classification Autoencoder (HCAE); and FFT in transformer-based model \cite{luo2024_ffttrans}, and 1DCNN \cite{zhang2023_fft1dcnn}. However, these models were developed for fault diagnosis of individual components and only considering single faults. On the other hand, Ding et al. \cite{ding2024_dataset} proposed a graph-based compound fault diagnosis model for multiple component system. However, it did not use any preprocessing and feature representation. In our work, we have both feature representation and fault diagnosis on multiple components with compound faults.

%\vspace*{0.05in} 

\section{Methodology}

In this section, we discuss the theoretical preliminaries for feature representation, data preprocessing including feature selection and normalization by speed working conditions, and deep learning architecture for feature extraction and classification.

\subsection{Preliminaries}

\subsubsection{Fourier Transform}

The digitalized form of the Fourier transform is called the discrete Fourier transform (DFT). Given a raw signal $x^{(c)}[n]$ from channel $c$, data points $N$, the DFT is defined as,

\vspace{-10pt}
\begin{equation}
\label{dft}
    X^{(c)}[k] = \sum^{N-1}_{n=0} x^{(c)}[n] e^{-j2\pi kn/N}
\end{equation}

for $k = 0,1,...,N-1$. 

The calculation of DFT can be sped up by considering its even and odd symmetries, this algorithm, called Fast Fourier Transform (FFT) \cite{cooley1965_fft}, is given by $X^{(c)}[k] = X^{(c)}_{even}[k] + X^{(c)}_{odd}[k]$, where,

\vspace{-10pt}
\begin{equation}
\label{fft_even}
    X^{(c)}_{even}[k] = \sum^{M-1}_{m=0} x^{(c)}[2m] e^{-j2\pi km/N}
\end{equation}

\vspace{-5pt}
\begin{equation}
\label{fft_odd}
    X^{(c)}_{odd}[k] = \sum^{M-1}_{m=0} x^{(c)}[2m+1] e^{-j2\pi k(m+1)/N}
\end{equation}

whereas $M = N/2$ and $m\in[0,M-1]$. Since the motor current and vibration signals are only real-valued, their FFTs are Hermitian symmetric. Thus, we only need to consider the real part of the magnitude spectrum for each channel $c$ given by,

\vspace{-10pt}
\begin{equation}
\label{fft_magnitude}
    |X^{(c)}[k]| = \sqrt{[\Re{(X^{(c)}[k]})]^2}
\end{equation}

% Include discussions on why the real part is only utilized

%\subsubsection{Fourier Transform Properties}
In addition to the simplified complexity due to the reduced input size, $N/2$, this magnitude spectrum $|X[k]|$ has several properties that are beneficial for feature representation:

\begin{itemize}
    \item \textit{Time Invariance.} FT is invariant to shifts in the time domain such that the FT of the shifted signal is the FT of the original signal with the same phase shift.

    \vspace{-10pt}
    \begin{equation}
    \label{time_invariant}
        \mathscr{F}(x[n]) = X[k] \Rightarrow \mathscr{F}(x[n-k]) = X[k]e^{-j2\pi kn/N}
    \end{equation}
    
    \item \textit{Magnitude Invariance.} If the signal is multiplied by a complex exponential, the magnitude $|X[k]|$ of the Fourier transform remains the same. This only causes phase shift in the amplitude spectrum that a CNN can capture.

    \vspace{-5pt}
    \begin{equation}
    \label{magnitude_invariant}
        \mathscr{F}(x[n]) = X[k] \Rightarrow \mathscr{F}(x[n]e^{j\phi}) = X[k]e^{j\phi}
    \end{equation}
    
    \item \textit{Linear Property.} The Fourier transform of a linear combination of signals is equivalent to the linear combination of their Fourier transforms. % This obeys the principle of superposition 

    \vspace{-5pt}
    \begin{equation}
    \label{linear_property}
        \mathscr{F}(ax_1[n]+bx_2[n]) = a |X_1[k]| + b |X_2[k]|
    \end{equation}
     %\mathscr{F}(ax_1[n]+bx_2[n]) = a \mathscr{F}(x_1[n]) + b \mathscr{F}(x_2[n])

    This property implies that the signals in frequency domain are viable inputs to the neural network since the neural network tries to derive the linear and non-linear relationships between the signals that would discriminate for each classes.
    
\end{itemize}

\subsubsection{Convolutional Neural Network}

CNNs \cite{lecun1995_cnn} have useful properties such as translational invariance, invariance to scaling if there is batch normalization, and invariance to local variations. The global average pooling averages the input feature map $B$ to have the sequence length reduced equivalent to kernel size $g$ given by,

\vspace{-15pt}
\begin{equation}
    \label{eq:avg_pool}
    G = \frac{1}{g} \sum^{g-1}_{i=0}{B_i}.
\end{equation}

Batch normalization normalizes the values using the mean $\mu_P$ and variance $\sigma^2_P$ of the mini batch $P$ with the trainable parameters $\gamma$ and $\beta$ pertaining to scaling and shifting respectively and a very small value $\epsilon$ to avoid division by zero.

\vspace{-5pt}
\begin{equation}
\label{eq:batchnorm}
    B_i = \gamma\cdot \frac{P_i-\mu_P}{\sqrt{\sigma^2_P+\epsilon}} + \beta
\end{equation}

Maximum pooling with the kernel size $p$ reduces the sequence length.

\vspace{-10pt}
\begin{equation}
\label{eq:maxpool}
    P_i = max(A_i,A_{i+1},...,A_{i+p+1})
\end{equation}

Applying rectified linear unit (ReLU) activation function on the Conv1D output adds non-linearity to the model using,

\vspace{-10pt}
\begin{equation}
\label{eq:relu}
    A_i = max(0,Y_i)
\end{equation}

Given the input sequence $x$, the output $Y$ of a one-dimensional convolution (Conv1D) of kernel size $k$ with trainable weights $w$ and bias $b$ is, 

\vspace{-10pt}
\begin{equation}
\label{eq:conv1d}
    Y_i = \sum^k_{j=1}{w_j \cdot x_{i+j-1}} + b.
\end{equation}

\subsubsection{Supervised Autoencoder}
\begin{figure}[tb]
    \centering
    \includegraphics[width=0.49\textwidth]{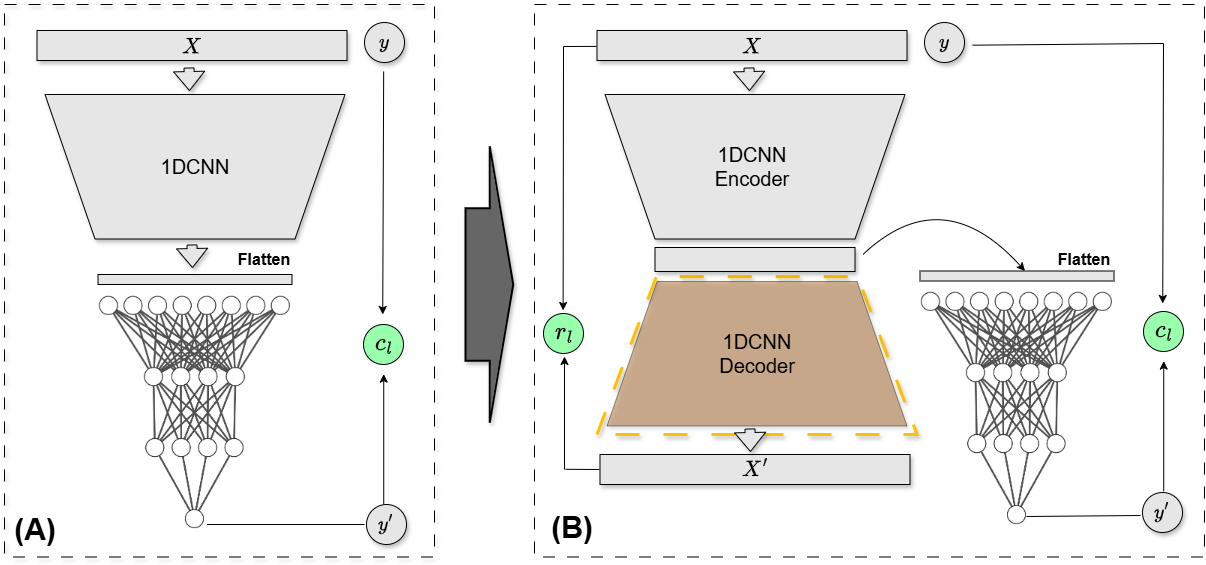}
    \caption{Illustration of (A) 1DCNN architecture to (B) supervised autoencoder architecture. Where $X$ and $y$ are the input data and binary label; while $X'$ and $y'$ are the reconstructed input data and the predicted label.}
    \label{fig:supervisedAE}
\end{figure}

The Supervised Autoencoder architecture introduced by \cite{le2018_sae}, is a regularizer similar to dropout layers, L1/L2 regularizer. The idea is that neural networks for supervised tasks can be further improved by incorporating an autoencoder architecture such that the loss function Eqn. (\ref{eq:supAE_loss}) becomes a sum of classification loss ($c_l$) and reconstruction loss ($r_l$). The architecture of supervised autoencoder from a 1DCNN architecture is illustrated in Fig. \ref{fig:supervisedAE}. The authors of [32] claim that the classification performance is never harmed due to the added regularizer. The training tends to be more stable because the model also learns the underlying structure of the data due to the added reconstruction loss.

\vspace{-10pt}
\begin{equation}
\label{eq:supAE_loss}
    loss = c_l + r_l
\end{equation}

\subsection{Data Preprocessing}

% Architecture

The data preprocessing as illustrated in the initial steps of Fig. \ref{fig:framework_fft-1dcnn} involves feature selection, amplitude spectrum calculation of the selected raw signals and data normalization according to speed working condition which is derived from a simple speed identification task (SIT). There are a total of 21 channels which can all be inputs to the model. However, not all of those channels are relevant for each component fault diagnosis. For example, vibration signals in the left axle box may not be relevant for motor fault diagnosis, since they are not directly connected in physical structure. Thus, we performed feature (channel) selection based on the physical structure and connections of the components. The motor channels CH1-CH9, the gearbox channels CH10-CH15, the left axle box channels CH16-CH18, and the right axle box channels CH19-CH21 are described in Table \ref{tab:channels}. The motor is connected to the gearbox, the left axle box is connected to the gearbox, and the right axle box is connected to the gearbox. Based on the selected raw input signals, we perform FFT to obtain the amplitude spectrum of each feature reducing the sequence length by half, from 64000 to 32000. The amplitude spectra are normalized according to the identified speed working condition, which is directly derived from the fundamental frequency, plus or minus some slip frequency, of its motor current signal CH7. The global minimum and maximum for each feature of the training set were stored for normalization of the validation and test sets. This benefits the training because neural network performs best when the input data are homogeneous and properly scaled. The normalized amplitude spectra are then fed into the 1DCNN model, which provides the binary classification of normal or anomaly for the particular fault type.

\begin{table}[tb]
\caption{Description of sensor channels.}
\centering
\label{tab:channels}
\begin{tabular}{lll}
\hline
Channel & Component & Signal Type \\ \hline
CH1, CH2, CH3 & Motor (drive end) & Tri-axial acceleration \\
CH4, CH5, CH6 & Motor (fan end) &  Tri-axial acceleration \\ 
CH7, CH8, CH9 & Motor (cable) & Three-phase current \\ 
CH10, CH11, CH12 & Gearbox (input axle) & Tri-axial acceleration \\ 
CH13, CH14, CH15 & Gearbox (input axle) & Tri-axial acceleration \\ 
CH16, CH17, CH18 & Axle box left (end cover) & Tri-axial acceleration \\ 
CH19, CH20, CH21 & Axle box right (end cover) & Tri-axial acceleration \\ \hline
\end{tabular}
\end{table}
\vspace{-5pt}

\subsection{Deep Learning}

\begin{figure*}[tb]
    \centering
    \includegraphics[width=0.8\textwidth]{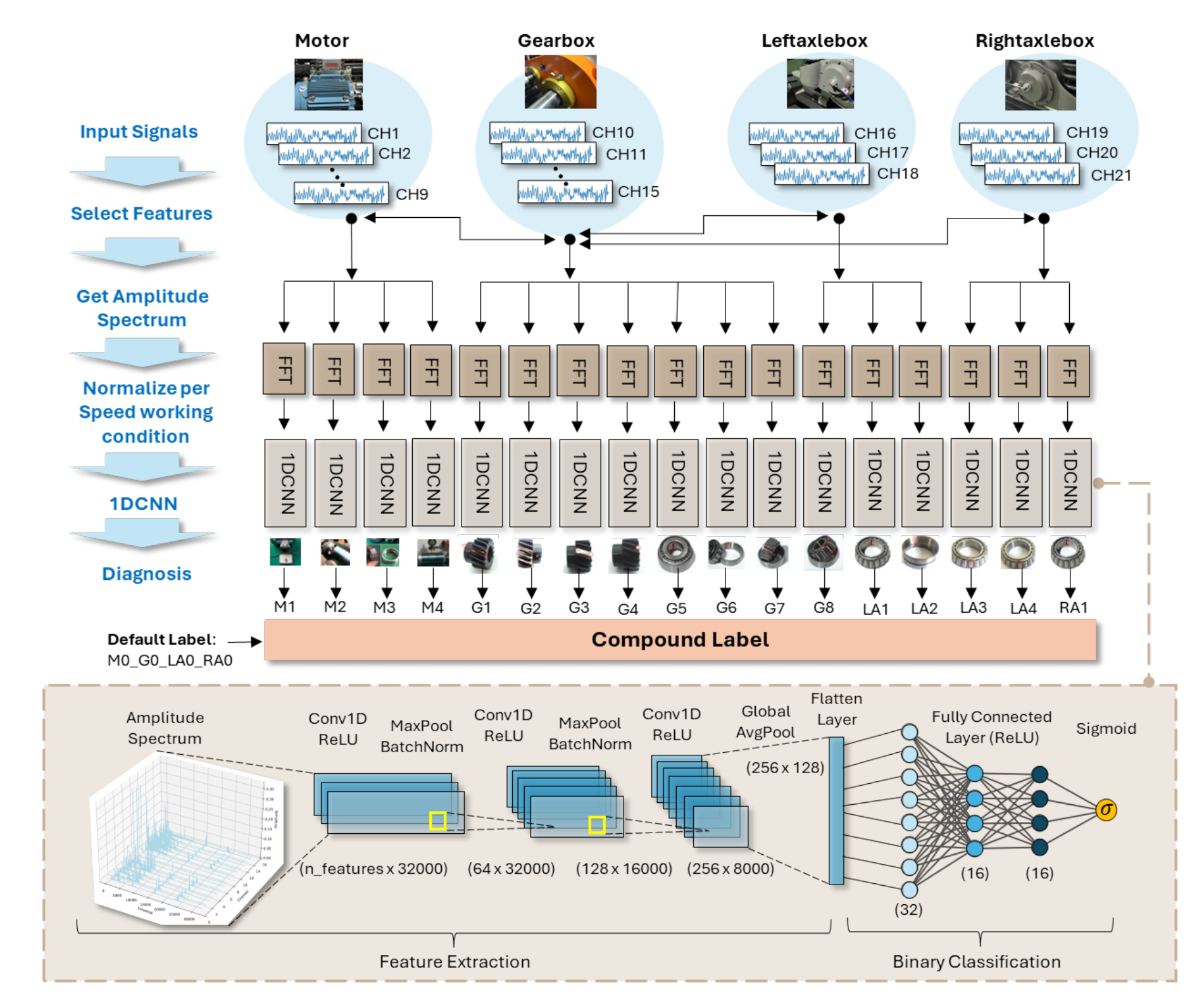}
    \caption{Proposed compound fault diagnosis framework using FFT-1DCNN.}
    \label{fig:framework_fft-1dcnn}
\end{figure*}

The multiclass multilabel classification was divided into multiple binary classification tasks equivalent to the number of single faults, i.e. 17 binary classification models. The proposed FFT-1DCNN framework illustrated in Fig. \ref{fig:framework_fft-1dcnn} is patterned before the 1DCNN architecture of \cite{ding2024_dataset} which includes convolutional layers with rectified linear unit (ReLU) activation functions, max pooling layers, batch normalization, and a global averaging layer. The convolutional layers have a kernel size of 9, with padding 1, and stride 1. The max pooling layers have a kernel size of 4 and stride of 2. The classifier neural network architecture is [32, 16, 16] and a sigmoid. 

%\begin{table}[htbp]
%\caption{Table Type Styles}
%\begin{center}
%\begin{tabular}{|c|c|c|c|}
%\hline
%\textbf{Table}&\multicolumn{3}{|c|}{\textbf{Table Column Head}} \\
%\cline{2-4} 
%\textbf{Head} & \textbf{\textit{Table column subhead}}& %\textbf{\textit{Subhead}}& \textbf{\textit{Subhead}} \\
%\hline
%copy& More table copy$^{\mathrm{a}}$& &  \\
%\hline
%\multicolumn{4}{l}{$^{\mathrm{a}}$Sample of a Table footnote.}
%\end{tabular}
%\label{tab1}
%\end{center}
%\end{table}

We compared our proposed FFT-1DCNN with a similar framework with supervised autoencoder architecture. Figure \ref{fig:supervisedAE} illustrates how it is implemented by adding a decoder network to the architecture so that the 1DCNN becomes the encoder. Then the flattened latent space also goes to the dense layer to obtain the binary classification. 

\section{Results and Discussion}

\subsection{Dataset Description}\label{dataset}
We use the publicly available train transmission dataset from the PHM-Beijing 2024 Data Challenge \cite{ding2024_dataset} which includes motor current and vibration signals of motor (M), gearbox (G), left axle (LA) box, and right axle (RA) box. Table \ref{tab:channels} shows the description and specific location of the sensors that measure the signals, also referred to as channels, CH1-CH21. This dataset considers the vibrational interactions between the components. There are several challenges in the dataset:

\begin{enumerate}
    \item \textbf{Multi-label classification.} Each sample can have normal or any combination of the other fault types of each component in Table \ref{tab:fault_types}. Such that for a sample, there are $(M)16\times(G)256\times(LA)16\times(RA)2= 131,072$ possible compound fault labels, calculated as the product of the combination of possible labels for each component, M, G, LA, and RA.

    %\begin{equation}
    %    \label{eq:count_labels}
    %    \# of possible\ labels=\left(M_{normal}+\sum M_{anomaly} \right) \times \left(G_{normal}+\sum G_{anomaly} \right)
    %    \times \left(LA_{normal}+\sum LA_{anomaly} \right)
    %    \times \left(RA_{normal}+\sum RA_{anomaly} \right)
    %\end{equation}
    
    \item \textbf{Compound faults.} Each label in the subway train transmission dataset can be a combination of faulty and normal components, or only faulty components, or only normal components. For example, the motor is normal but the other components are faulty, in this case, the label is $M0\_G1\_LA2\_RA1$. In such cases, the vibration signals might be affected by the fault of other components. Table \ref{tab:compound_labels} shows the 42 compound faults in this dataset.
    
    \item \textbf{Varying and unknown working conditions} Signal data were collected from the subway train transmission system with 9 different working conditions, combination of different speed $v\in[20Hz, 40Hz, 60Hz]$ and lateral load $l\in[-10 kN, 0 kN, 10 kN]$ working conditions. In addition, the dataset has information on the working conditions only for preliminary training and test sets. The rest of the samples do not have information on the working conditions.
    
    \item \textbf{Imbalanced data.} The initial training set consists of 3 samples for each single fault type. For most fault types, when the data is split for binary classification, the normal to anomaly ratio is about 20:1 in most cases of the training set, as shown in Table \ref{tab:data_splitting}. 
\end{enumerate}

\begin{table}[tb]
\caption{Description of fault types.}
\centering
\label{tab:fault_types}
\begin{tabular}{cl}
\hline
Fault Code & Description \\ \hline
M0,G0,LA0,RA0 & Normal component \\
M1 &  Motor - short circuit \\ 
M2 &  Motor - broken rotor bar \\ 
M3 &  Motor - bearing fault \\ 
M4 &  Motor - bowed rotor \\ 
G1 &  Gearbox - gear cracked tooth \\ 
G2 &  Gearbox - gear worn tooth \\ 
G3 &  Gearbox - gear missing tooth \\ 
G4 &  Gearbox - gear chipped tooth \\ 
G5 &  Gearbox - bearing inner race fault \\ 
G6 &  Gearbox - bearing outer race fault \\ 
G7 &  Gearbox - bearing rolling element fault \\ 
G8 &  Gearbox - bearing cage fault \\ 
LA1 &  Left Axle Box - bearing inner race fault \\ 
LA2 &  Left Axle Box - bearing outer race fault \\ 
LA3 &  Left Axle Box - bearing rolling element fault \\ 
LA4 &  Left Axle Box - bearing cage fault \\ 
RA1 &  Right Axle Box - bearing inner race fault\\ \hline
\end{tabular}
\end{table}

\subsection{Evaluation Metrics}

The main performance metric for model evaluation is the Z metric which is a weighted combination of accuracy, precision, recall, and f1 score introduced in the PHM-Beijing 2024 Data Challenge. This metric provides a better evaluation than the commonly used metric, accuracy, which does not provide a clear indication if the data is highly imbalanced. In the discussions, we often refer to accuracy to compare with existing studies, since the Z metric was only introduced in 2024 for the PHM Data Challenge, \href{https://2024.icphm.org/datachallenge/}{2024.icphm.org}.

\vspace{-10pt}
\begin{equation}
    Z = 0.4(Acc) + 0.2(Prec) + 0.2(Rec) + 0.2(F1)
\label{eqn:Z_metric}
\end{equation}

where,
\vspace{-5pt}
\begin{equation}
    Accuracy = \frac{TP + TN}{TP+TN+FP+FN}
\label{eqn:accuracy}
\end{equation}

\begin{equation}
    Precision = \frac{TP}{TP+FP}
\label{eqn:precision}
\end{equation}

\begin{equation}
    Recall = \frac{TP}{TP+FN}
\label{eqn:recall}
\end{equation}

\begin{equation}
    F1 = 2\times\frac{Precision\times Recall}{Precision+Recall}
\label{eqn:f1}
\end{equation}

Where TP, TN, FP, and FN are true positives, true negatives, false positives, and false negatives, respectively.

Moreover, we measured the model complexity by calculating the floating point operations per second (FLOPs) of each model, exclusive of the preprocessing steps, using Pytorch \textit{ptflops} module. A higher FLOPs metric means that the model is more complex and has a longer run time.

\begin{table}[tb]
    \centering
    \caption{Forty two (42) compound labels of PHM Beijing 2024 Dataset and whether they are present in training set, preliminary test set, and final test set.}
    \label{tab:compound_labels}
    \begin{tabular}{llccc}
    \hline
       No. & Compound label & Train & \multicolumn{2}{c}{Test}\\
        \cline{4-5}
        & & & Prelim & Final \\ \hline
       1 & M0\_G0\_LA0\_RA0 & \checkmark & \checkmark & \checkmark\\
       2 & M1\_G0\_LA0\_RA0 & \checkmark & \checkmark & \checkmark\\
       3 & M2\_G0\_LA0\_RA0 & \checkmark & \checkmark & \checkmark\\
       4 & M3\_G0\_LA0\_RA0 & \checkmark & \checkmark & \checkmark\\
       5 & M4\_G0\_LA0\_RA0 & \checkmark & \checkmark & \checkmark\\
       6 & M0\_G1\_LA0\_RA0 & \checkmark & \checkmark & \checkmark\\
       7 & M0\_G2\_LA0\_RA0 & \checkmark & \checkmark & \checkmark\\
       8 & M0\_G3\_LA0\_RA0 & \checkmark & \checkmark & \checkmark\\
       9 & M0\_G4\_LA0\_RA0 & \checkmark & \checkmark & \checkmark\\
       10 & M0\_G5\_LA0\_RA0 & \checkmark & \checkmark & \checkmark\\
       11 & M0\_G6\_LA0\_RA0 & \checkmark & \checkmark & \checkmark\\
       12 & M0\_G7\_LA0\_RA0 & \checkmark & \checkmark & \checkmark\\
       13 & M0\_G8\_LA0\_RA0 & \checkmark & \checkmark & \checkmark\\
       14 & M0\_G0\_LA1\_RA0 & \checkmark & \checkmark & \checkmark\\
       15 & M0\_G0\_LA2\_RA0 & \checkmark & \checkmark & \checkmark\\
       16 & M0\_G0\_LA3\_RA0 & \checkmark & \checkmark & \checkmark\\
       17 & M0\_G0\_LA4\_RA0 & \checkmark & \checkmark & \checkmark\\
       18 & M0\_G0\_LA1+LA2+LA4\_RA0 & \checkmark &  & \checkmark\\
       19 & M0\_G4+G5\_LA0\_RA0 & \checkmark &  & \checkmark\\
       20 & M1\_G0\_LA1\_RA1 & \checkmark &  & \checkmark\\
       21 & M0\_G3\_LA1\_RA0 & \checkmark &  & \checkmark\\
       22 & M1\_G0\_LA1\_RA0 & \checkmark &  & \checkmark\\
       23 & M4\_G3\_LA0\_RA0 &  &  & \checkmark\\
       24 & M0\_G1+G5\_LA0\_RA0 &  &  & \checkmark\\
       25 & M0\_G0\_LA2+LA3\_RA0 &  &  & \checkmark\\
       26 & M2\_G0\_LA1\_RA0 &  &  & \checkmark\\
       27 & M0\_G0\_LA2+LA4\_RA0 &  &  & \checkmark\\
       28 & M3\_G3\_LA0\_RA0 &  &  & \checkmark\\
       29 & M1\_G5\_LA0\_RA0 &  &  & \checkmark\\
       30 & M0\_G2+G5\_LA0\_RA0 &  &  & \checkmark\\
       31 & M0\_G0\_LA1+LA2\_RA0 &  &  & \checkmark\\
       32 & M1\_G3\_LA0\_RA0 &  &  & \checkmark\\
       33 & M3\_G0\_LA1\_RA0 &  &  & \checkmark\\
       34 & M3\_G5\_LA0\_RA0 &  &  & \checkmark\\
       35 & M0\_G0\_LA1\_RA1 &  &  & \checkmark\\
       36 & M0\_G3+G5\_LA0\_RA0 &  &  & \checkmark\\
       37 & M0\_G0\_LA1+LA2+LA3+LA4\_RA0 &  &  & \checkmark\\
       38 & M0\_G0\_LA1+LA2+LA3\_RA0 &  &  & \checkmark\\
       39 & M2\_G5\_LA0\_RA0 &  &  & \checkmark\\
       40 & M4\_G5\_LA0\_RA0 &  &  & \checkmark\\
       41 & M2\_G3\_LA0\_RA0 &  &  & \checkmark\\
       42 & M2\_G0\_LA1\_RA1 &  &  & \checkmark\\ \hline

    \end{tabular}
\end{table}
\vspace{-5pt}

\subsection{Implementation Details}
The numerical simulation was implemented using Python language and Pytorch deep learning library on a computer workstation with a NVIDIA 16 GB GPU GeForce RTX 4080 SUPER. The training data were divided into 1-second nonoverlapping time slices, that is, 64000 timesteps per slice given that the sampling rate is 64 kHz. The training set is further divided into roughly 80\% train and 20\% validation sets depending on the fault type. Table \ref{tab:data_splitting} shows the exact number of samples in each set after data splitting. There are two holdout test sets: the preliminary test set contains 102 samples of single faults, and the final test set contains 252 samples of both single and compound faults. The initial learning rates for training the 1DCNN is 0.0001 and 0.001 for the dense layer. The loss function is weighted binary cross entropy, \textit{BCEwithLogitsLoss} and the gradient descent optimizer is the Adam optimizer. The models are trained for 100 iterations and a batch size of 32. The model weights of the epoch with the lowest validation loss were further used to test the model performance.

\begin{comment}
The preliminary test set is excluded in Table \ref{tab:data_splitting} because the number of samples for each fault type in preliminary test set are all 106 normal and 6 anomaly, except for RA1 fault where there are 112 normal and 0 anomaly.
\end{comment}

\begin{table}[tb]
    \centering
    \caption{Data splitting for training, validation, and hold out final test set with compound faults.}
    \label{tab:data_splitting}
    \begin{tabular}{lcccccc}
    \hline
        Fault & \multicolumn{2}{c}{Train} & \multicolumn{2}{c}{Validation} & \multicolumn{2}{c}{Test} \\
        \cline{2-7} 
         & \footnotesize normal & \footnotesize anomaly & \footnotesize normal & \footnotesize anomaly & \footnotesize normal & \footnotesize anomaly  \\ \hline
        M1 & 815 & 75 & 211 & 21 & 222 & 30 \\
        M2 & 840 & 40 & 222	& 20 & 222 & 30 \\
        M3 & 840 & 40 & 222	& 20 & 228 & 26 \\
        M4 & 840 & 40 & 222	& 20 & 234 & 18\\
        G1 & 840 & 40 & 222	& 20 & 240 & 12\\
        G2 & 840 & 40 & 222	& 20 & 240 & 12\\
        G3 & 835 & 45 & 221	& 21 & 210 & 42\\
        G4 & 829 & 70 & 203	& 20 & 240 & 12\\
        G5 & 829 & 70 & 203	& 20 & 198 & 54\\
        G6 & 840 & 40 & 222	& 20 & 246 & 6\\
        G7 & 840 & 40 & 222	& 20 & 246 & 6\\
        G8 & 840 & 40 & 222	& 20 & 246 & 6\\
        LA1 & 790 & 100 & 200 & 32 & 180 & 72\\
        LA2 & 838 & 70 & 194 & 20 & 210	& 42\\
        LA3 & 820 & 70 & 212 & 20 & 228	& 24\\
        LA4 & 838 & 70 & 194 & 20 & 228	& 24\\
        RA1 & 894 & 4 & 222	& 2 & 234 & 18\\
    \hline
    \end{tabular}
\end{table}
\vspace{200pt}

\subsection{Feature Representation}

\begin{figure}[tb]
    \centering
    \includegraphics[width=0.5\textwidth]{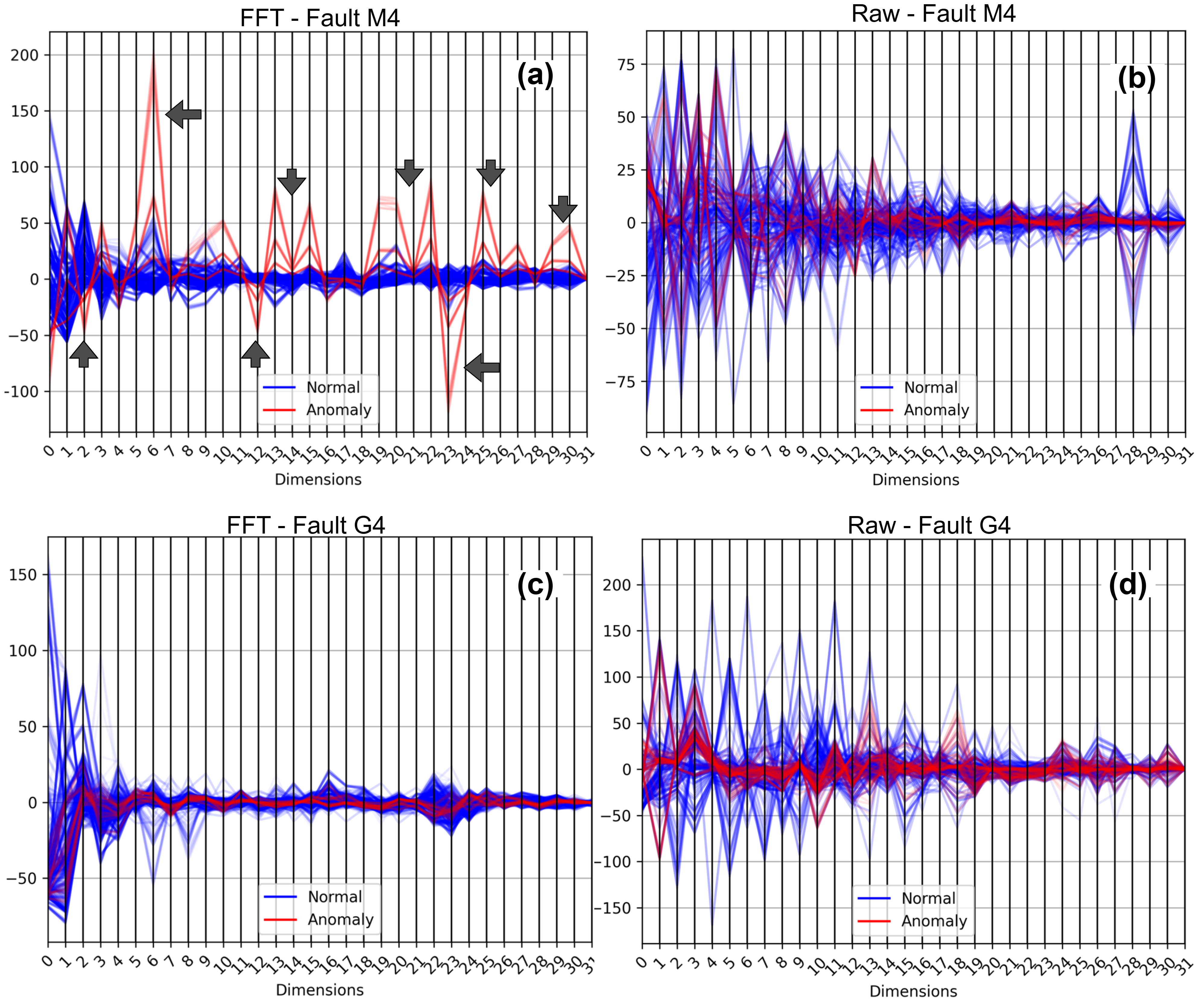}
    \caption{Parallel coordinates plots of 1DCNN output feature space of the best case scenario (M4 fault) (a) with FFT representation, (b) using raw signals only, and the worst case scenario (G4 fault) (c) with FFT representation, (d) using raw signals only. The arrows in (a) indicate feature dimensions where there are clear disentanglement between classes. On the other hand, (b), (c), and (d) show that most feature dimensions are entangled.}
    \label{fig:feature_space}
\end{figure}

The feature space representation is visualized by plotting the parallel coordinates of the feature space output of the 1DCNN. Unlike t-SNE and UMAP which requires data reduction down to 2 or 3 dimensions, parallel coordinates plot can be visualized even with high dimension, e.g. 32 dimensions. From the latent space of size $256\times128$, we used the top 32 eigenvalues based on the principal component analysis (PCA). Fig. \ref{fig:feature_space} shows parallel coordinate plots for 32 dimensions of different classification performance scenarios, such as fault M4 and fault G4. From this figure, we can show that the Fourier transform helps improve the feature representation of the data for fault diagnosis. For example, Fig. \ref{fig:feature_space}a shows several class separations in various coordinates. However, it is difficult to observe the separations from the rest of the plots, Fig. \ref{fig:feature_space}b-d. 

\vspace{-5pt}

\subsection{Ablation Study}

We performed ablation experiments on the proposed framework to examine the impact of each preprocessing step on the model performance. That is, by removing some of the steps or procedures in the algorithm, we analyze the effect of the performance. The first ablation experiment is performed by Raw-1DCNN without the Fourier transform preprocessing and the accuracy in the final test set was reduced by 17\%. The second ablation experiment skipped the feature selection, FFT-1DCNN (all features), which slightly reduced the accuracy in the final test set by 1.3\%. The results of the ablation experiments are summarized in Table \ref{tab:ablation_study}.

\begin{table}[tb]
\caption{Ablation study results.}
\centering
\label{tab:ablation_study}
\begin{tabular}{lcccc}
\hline
Model & \multicolumn{2}{c}{Accuracy} & \multicolumn{2}{c}{Z} \\ 
\cline{2-3} \cline{4-5}
& Prelim & Final & Prelim & Final \\ \hline
Raw-1DCNN & 0.7750 & 0.7617 & 0.6251 & 0.5972\\
FFT-1DCNN (all features) & 0.9725 & 0.9263 & 0.9460 & 0.8614\\
FFT-1DCNN & \textbf{0.9767} & \textbf{0.9393} & \textbf{0.9543} & \textbf{0.8866}\\ \hline
\end{tabular}
\end{table}

\subsection{Classification}

Table \ref{tab:benchmark_performance} shows the comparison between the traditional unsupervised autoencoder, supervised convolutional autoencoder, and 1DCNN. It shows that the traditional unsupervised autoencoder, even with convolutional layers, does not perform well in this dataset's classification task. This is because the autoencoders prioritize learning the underlying structure by reconstructing the input data. On the other hand, by incorporating an additional regularizer into the FFT-1DCNN model using the supervised autoencoder architecture,  its training stability and generalization improved by an accuracy of 93.97\% on the final test set. The loss function is shown in Eqn. (\ref{eq:supconvAE_loss}) which is a sum of binary cross entropy (BCE) and mean squared error (MSE). The constants $\lambda_0$ and $\lambda_1$ refer to the sensitivities of the BCE and MSE loss respectively. In this study, we have set $\lambda_0=1.0$ and $\lambda_1=0.1$ as the best performing sensitivity hyperparameters from the search space $\lambda_0 \in [0.01, 0.1, 1.0]$ and $\lambda_1\in [0.01, 0.1, 1.0]$. Although the supervised autoencoder improved the performance, the improvement (+0.04\% accuracy) is still comparable to the 1DCNN model and the supervised autoencoder has increased the model complexity by 6.8 MFLOPs from that of 1DCNN.

\vspace{-5pt}
\begin{equation}
    \label{eq:supconvAE_loss}
    loss = \lambda_{0}\ BCE + \lambda_{1}\ MSE
\end{equation}

\begin{table}[tb]
\caption{Summary of performance on benchmark models.}
\centering
\label{tab:benchmark_performance}
\begin{tabular}{lccccc}
\hline
Model & \multicolumn{2}{c}{Accuracy} & \multicolumn{2}{c}{Z} & FLOPs \\ 
\cline{2-3} \cline{4-5}
& Prelim & Final & Prelim & Final & \\ \hline
FFT-UnsupAE & 0.5691 & 0.5480 & 0.4046 & 0.4075 & 11.9 M \\
FFT-SupConvAE & 0.9757 & \textbf{0.9397} & 0.9526 & \textbf{0.8871} & 30.6 M\\
FFT-1DCNN & \textbf{0.9767} & 0.9393 & \textbf{0.9543} & 0.8866 & 23.8 M\\ \hline
\end{tabular}
\end{table}

\begin{figure}[tb]
    \centering
    \includegraphics[width=0.45\textwidth] {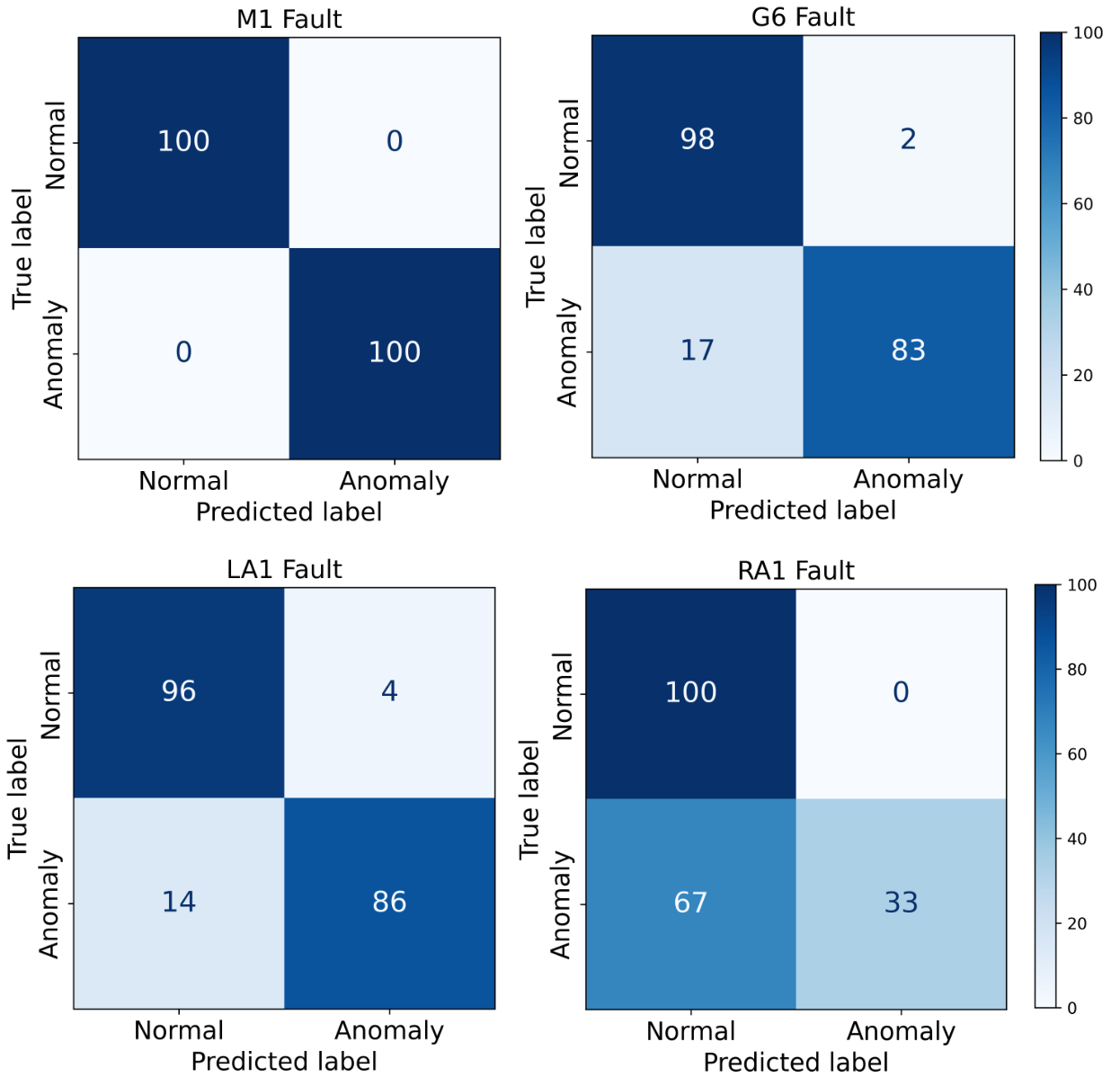}
    \caption{Confusion matrices of selected faults on the final test set using FFT-1DCNN. The values are shown as percentages (\%) due to data imbalance between classes.}
    \label{fig:cm_motor}
\end{figure}

Our model achieved 93.93\% and 97.67\% (binary class) accuracies and compared to the state-of-the-art, \cite{ding2024_dataset} achieved 94\% to 98\% diagnostic multiclass accuracies in a similar compound fault diagnosis dataset but different data settings. Fig. \ref{fig:cm_motor} shows the confusion matrices for selected fault types M1, G6, LA1, and RA1. In terms of model complexity, \cite{ding2024_dataset} has 4.7 GFLOPs while our proposed model only has 23.8 MFLOPs.

\section{Conclusion}

In this paper, we investigate the advantages of using Fourier transform for time series data representation of vibration and motor current signals of subway train transmission system. We developed the FFT-1DCNN framework for subway train fault diagnosis using the recent PHM-Beijing 2024 dataset. We have shown that the frequency domain provides a better data representation as input for the fault diagnosis model utilizing a 1DCNN architecture, based on the model performance and parallel coordinates plots of the feature space. Our model has a comparable accuracy to that of FFT-SupervisedConvAE and \cite{ding2024_dataset} while having a lower model complexity in terms of FLOPs.

% key takeaways
% Fourier representation
% feature selection for interacting components
% speed working condition normalization
% 

Our proposed framework has trade-offs in terms of model complexity and the number of models. The model's simple architecture of only 23.8 MFLOPs is also composed of several models (17 binary classifiers), which might be difficult to train and manage for industrial applications. The model also performs worse in gearbox faults, since all the other 3 components are directly connected to the gearbox, affecting its vibration signals. Moreover, the model does not perform well in the RA1 fault due to the small number of samples on this fault type. Lastly, the model assumes that the input data have high sampling rates, e.g. 64kHz, and from multiple sensor channels attached close to each transmission component. This is because FFT suffers from aliasing and spectral leakage if the sampling rate is low and the model may not be able to distinguish between vibrations from different components if the sensor is only attached to the housing.

% hypothesis
% assumptions
% limitations
% all negative attributes of the paper

Our future work involves incorporating an evolvable CNN architecture \cite{aradhya2022_autocnn} into our model framework for continual learning tasks. Such a model will be able to adapt to concept drifts, i.e. change of data distribution, while minimizing catastrophic forgetting. Furthermore, the study can be extended to include the enhanced FFT \cite{lin2016_efft} which emphasizes the frequency peaks by minimizing the effects of spectral leakage.

% multiclass classification
% graph neural network
% enhanced Fourier 

\section*{Declaration of Competing Interest}
The authors declare that they have no known competing financial interests or personal relationships that could have appeared to influence the work reported in this paper.

\section*{Data availability}
The PHM Beijing 2024 dataset \cite{ding2024_dataset} is publicly available and can be accessed through their official website, \href{https://2024.icphm.org/datachallenge/}{2024.icphm.org}. 

\section*{Acknowledgment}

The first author acknowledges Singapore University of Technology and Design (SUTD) and the Ministry of Education, Singapore (MOE) for the Research Student Scholarship (RSS) and SUTD Ph.D. Fellowship for his doctoral studies from 2023-2026. The logistics of this research was also supported by the Research Surplus Grant No. RGSUR08 of Prof. Nagarajan Raghavan. The authors would also like to thank the Agency for Science, Technology, and Research (A*STAR) for providing computational resources for this study.

%\section*{References}

\begin{comment}

\clearpage
\appendix

\begin{figure*}[tb]
    \centering
    \includegraphics[width=0.9\textwidth]{figures/train_val_loss_fft-1dcnn.pdf}
    \caption{Training and validation loss per iteration of FFT-1DCNN on each fault type.}
    \label{fig:train_val_loss_fft-1dcnn}
\end{figure*}

%\begin{figure*}[tb]
    %\centering
    %\includegraphics[width=\textwidth]{figures/train_val_loss_SupConvAE.pdf}
    %\caption{Training and validation loss per iteration for FFT-SupervisedConvAE.}
    %\label{fig:train_val_loss_SupConvAE}
%\end{figure*}

\begin{figure*}[tb]
    \centering
    \includegraphics[width=0.9\textwidth]{figures/confusion_matrix_fft-1dcnn.pdf}
    \caption{Confusion matrices of FFT-1DCNN on each fault type.}
    \label{fig:confusion_matrix_fft-1dcnn}
\end{figure*}

%\begin{figure*}[tb]
%    \centering
    %\includegraphics[width=\textwidth]{figures/confusion_matrix_SupConvAE.pdf}
    %\caption{Confusion matrices for FFT-SupervisedConvAE.}
    %\label{fig:confusion_matrix_SupConvAE}
%\end{figure*}

\end{comment}

\end{document}